# Data-Driven Pixel Control: Challenges and Prospects

Saurabh Farkya, Zachary Alan Daniels, Aswin Raghavan, Gooitzen van der Wal, Michael Isnardi, Michael Piacentino, and David Zhang

Center for Vision Technologies, SRI International, Princeton NJ 08540, USA
{saurabh.farkya, zachary.daniels, aswin.raghavan, gooitzen.vanderwal, michael.isnardi, michael.piacentino, david.zhang}@sri.com

**Abstract.** Recent advancements in sensors have led to high resolution and high data throughput at the pixel level. Simultaneously, the adoption of increasingly large (deep) neural networks (NNs) has lead to significant progress in computer vision. Currently, visual intelligence comes at increasingly high computational complexity, energy, and latency. We study a data-driven system that combines dynamic sensing at the pixel level with computer vision analytics at the video level and propose a feedback control loop to minimize data movement between the sensor front-end and computational back-end without compromising detection and tracking precision. Our contributions are threefold: (1) We introduce *anticipatory attention* and show that it leads to high precision prediction with sparse activation of pixels; (2) Leveraging the feedback control, we show that the dimensionality of learned feature vectors can be significantly reduced with increased sparsity; and (3) We emulate analog design choices (such as varying RGB or Bayer pixel format and analog noise) and study their impact on the key metrics of the data-driven system. Comparative analysis with traditional pixel and deep learning models shows significant performance enhancements. Our system achieves a 10X reduction in bandwidth and a 15-30X improvement in Energy-Delay Product (EDP) when activating only 30% of pixels, with a minor reduction in object detection and tracking precision. Based on analog emulation, our system can achieve a throughput of 205 megapixels/sec (MP/s) with a power consumption of only 110 mW per MP, i.e., a theoretical improvement of ∼30X in EDP.

## 1 Introduction

Imaging systems are generating increasingly large amounts of data, driven by advancements in sensor technology and algorithms requiring higher frame rate data collection, high dynamic range data, and continuous uninterrupted surveillance. Edge devices, constrained by size, weight, and power (SWaP), struggle to process high-bandwidth data streams with low latency. We propose leveraging data-driven learning to reduce data bandwidth, energy usage, and latency on edge hardware, enabling effective real-time analysis of complex scenes.

Our system follows the Dynamic Data Driven Applications Systems (DDDAS) paradigm [11], which tightly integrates an architecture for dynamic sensing with intelligent data processing in a feedback control loop. Not all pixels in a scene are equally important [2,3,17,19]. By processing a subset of *salient* pixels [4,16], one can improve the efficiency and accuracy of machine learning (ML) models for computer vision (CV). Inspired by *saccades* in biological vision (quick eye movements between multiple fixation points influenced by attention [22]), we explore *anticipatory sensing* [15,28], where task-oriented visual saliency is predicted for future frames using earlier outputs.

---

[0] Demo for the full end-to-end system design: Link to demo
[1] Subsampled MSCOCO available here.
[2] **Distribution Statement A.** Approved for Public Release: Distribution is Unlimited.

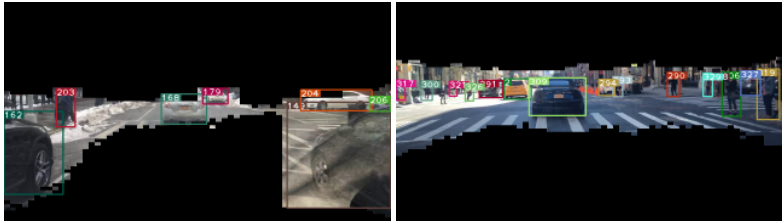

**Fig. 1.** Object tracking in video with pixels selected using our *anticipatory attention*
.

The proposed system can leverage any imager that supports two core functionalities: 1) dynamic activation (turning on/off) of a patch of pixels in the sensor array and analog-digital conversion only on the activated pixels and 2) simple feature extraction (e.g., linear operations) on the activated patches (e.g., our sensor [39]). Pixel-level feature extraction is more feasible when few patches are active due to shared analog circuitry.

At every time step, the front-end sensor extracts features from a few patches in an image (see Fig. 1), digitizes the features, and feeds them into a back-end deep learning (DL) model. The DL model uses this information to update internal states, perform visual understanding (e.g., detection/tracking), and anticipate patches to activate in the next time step to maximize task-level performance within the constraints of the imager.

Our method reduces power consumption by only activating patches that are digitized. While we only consider binary throttling of patches in this paper, fine-grained control of patch-level imager characteristics (e.g., exposure) is also possible. Different hardware constraints and impact on power consumption should be explored in future work. Intuitively, sparse sensor data from the front-end can be processed by a less complex DL back-end (e.g., smaller models, fewer bits). A complex back-end for predicting the next time-step patches can introduce latency. We propose Energy-Delay Product (EDP) as a key metric to quantify the interplay between the number of patches and back-end model complexity. EDP captures the impact of design choices in both the front-end and back-end. Our trade studies compare EDP with object tracking precision.

Our work shares similarities to other models that combine active sensing with DL models, such as Deep Anticipatory Networks [28], Glance and Focus Networks [21], shift-aware models for visual saliency object detection [13], glimpse prediction [27], and deep reinforcement learning approaches for learning how to pan, zoom, and/or focus [1,5,26,32,33,40]. In contrast to the prior DDAS approaches (e.g., active scene classification [9,10] and object tracking [30,31,35,36]), this work focuses on *the anticipatory* prediction of salient patches and exploits dynamic sensing for the purposes of reducing bandwidth, energy, and latency. The novel contributions of this paper are:

- A framework for data-driven learning of sparse anticipatory attention to maximize precision on downstream scene understanding tasks within imager constraints.

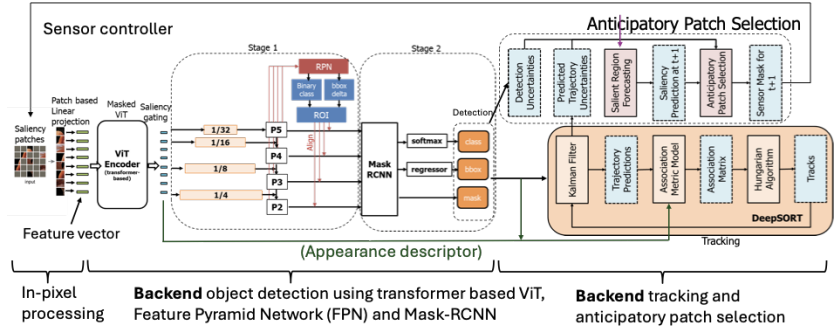

**Fig. 2.** An overview of the system architecture from the front-end sensing to back-end prediction

- Techniques for DL model compression that leverage the sparsity in the sensor data to reduce the dimensionality of feature vectors and increase their sparsity.
- A system solution for multi-object tracking in video, where anticipatory sensing combined with in-pixel processing leads to a $\sim$10X reduction in bandwidth and a 15-30X improvement in energy-latency savings compared to traditional hardware.
- An empirical robustness study of the proposed system with emulation of the underlying analog design choices (pixel design, analog noise) on several benchmark datasets (e.g., MS-COCO, MOT17, and BDD100K).

## 2    Technical Approach

### 2.1    Overview of Proposed Framework

We aim to move processing into the sensor pixel to avoid power consumption due to data movement. In prior work [39], we presented an analog design that computes salient patch features before analog-to-digital conversion. Our framework for dynamic scene understanding is versatile, but this paper's focus is Multiple Object Tracking (MOT) [8]. The system architecture (Fig. 2) illustrates the process from front-end sensing through back-end prediction to feedback control. The anticipatory sensing mechanism tells the sensor which patches to sense. The in-pixel processor uses patch-level linear projections to extract basic features for the sensed patches. By only processing a subset of patches and compressing this information into feature vectors, the front-end reduces the amount of data sent to the back-end. These features are sent to a Vision Transformer (ViT) [12], which uses self-attention [34] to process the partially sensed image, aggregating and transforming the features for downstream tasks. The ViT features are fed into a Feature Pyramid Network (FPN) [24] and Mask-RCNN [18] to perform object detection. The object detections along with ViT features feed into a DeepSORT model [37] to perform MOT. The object detector features are re-used with a recurrent neural network (RNN) [7] to forecast which patches will contain salient objects in future frames. RNN-based saliency scores, detection uncertainties, and tracking uncertainties are ensembled to prioritize patch sensing for the next frame, promoting efficient identification of new objects and minimizing the uncertainty of existing object tracks. Every $N = 16$ frames, our system fully senses the image to avoid missing new objects.

## 2.2    Masked Autoencoder and Object Detection

Mask-RCNN [18] serves as the object detector. To extend the model's feature extractor for processing partial information, we adapt the randomized masking technique from Masked ViT [19], which randomly drops patches from the input image during training and forces the model to reconstruct the missing information. By dropping $p\%$ of the total image patches $n$ (features), the computational complexity of the ViT's attention mechanism is reduced to $\mathscr{O}((p*n)^2), p < 1$. For further improvement, we move from random to targeted salient masking.

In conventional detectors, the FPN [24] is commonly used to facilitate detection across different scales. Fang et al. [14] extend ViT and FPN to work with masked models for detection but require expensive fine-tuning of the pre-trained ViT model. Our design avoids retraining by leveraging features extracted from various layers of the ViT encoder, reconstructing missing features, and integrating with Mask-RCNN. Our current approach doesn't fully leverage partial information post-FPN, but future work could utilize models like DeTR [6] to overcome this limitation.

We utilize the Masked ViT model as a fixed feature extractor while fine-tuning the overall model using the Mask-RCNN objective function. During training, we mask $\geq 70\%$ of patches within images, extracting features from $\sim 30\%$, mimicking our anticipatory sensing framework for MOT. This method reduces the region-of-interest for the detector, streamlining object detection (see Table 1). The anticipatory mechanism may not always provide a perfect set of patches. To fortify the detector, during training we introduce noise by: 1) randomly excluding patches inside large objects and 2) including a few non-salient background patches.

## 2.3    Object Tracking

The MOT model consists of a straightforward implementation of DeepSORT. SORT uses Kalman filtering to predict object trajectories in image space and associate objects via the Hungarian algorithm using bounding box overlap. DeepSORT improves association using visual feature matching (in our system, using ViT encoder features).

## 2.4    Anticipatory Sensing

The tracking model predicts where an object will appear in the next frame. Our anticipatory attention mechanism scores the importance of each object for the next time frame to rapidly identify new objects and reduce uncertainty in existing detections/tracks:

– **Predicted saliency**: Using global image features, we train an RNN to predict which patches will contain salient objects in the next frame using ground truth segmentations during training. We assign a probability of saliency to each patch and compute an object-level score by averaging over all patches overlapping the object.
– **Detection uncertainties**: We use one minus the confidence output by the detector as a score, forcing the system to focus on objects with high detection uncertainty.
– **Tracking uncertainties**: The Kalman filter of DeepSORT outputs covariance matrices representing uncertainty in each object's position. We measure the change in the size of the covariance matrix from the previous to the current time step $t$: $\max(\frac{\det(cov(obj,t))-\det(cov(obj,t-1))}{\det(cov(obj,t))+\varepsilon}, 0)$. The system should focus on objects where by not sensing the object, the uncertainty in the object's position grows.

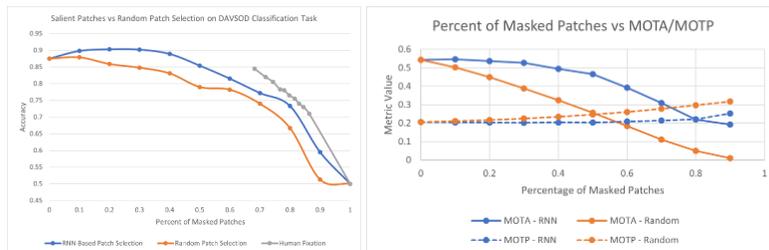

**Fig. 3.** Left: Comparison of learned vs human-driven saccades on the DAVSOD dataset. Human fixation data points are determined by thresholding different percentages of salient pixels per patch. Right: Comparing the RNN-based model with random patch selection on MOT-17.

These scores are combined with equal weight for a final object-level score. Objects are sorted by score, and 30% of patches that overlap with the highest-scoring objects are selected. If < 30% of patches are selected, the RNN selects the remaining patches.

### 2.5  Model Compression

To optimize our model for deployment on edge devices, we employ a two-stage approach. In stage 1, we downsized ViT from 768 dimensions to 240 dimensions by applying knowledge distillation [20]. This process involves using the $\ell 2$ loss across the entire FPN to ensure a smooth convergence while maintaining the same Mask R-CNN model between the 768D and 240D ViT models. In stage 2, with the FPN reduced to 30% of its original size from stage 1, we further decrease the dimensions of all Mask R-CNN modules by 25%, and fine-tune the model using the Mask R-CNN loss function.

## 3  Experimental Results

### 3.1  Importance of Salient Patches

Our first set of experiments focuses on understanding the importance of salient patch selection. Additional studies can be found in Farkya et al. [15]. In these experiments, we do not utilize the full system; instead, we train an RNN to predict salient patches in future frames and use off-the-shelf task-specific models on partially sensed images. We evaluate object recognition from video clips using the DAVSOD dataset [13], which contains videos annotated with human eye fixations. We consider: 1) random patch selection, 2) oracle-based patch selection (determined by human fixation), and 3) RNN-based patch selection. The evaluation uses per-frame classification accuracy over four classes (human, animal, artifact, and vehicle). The RNN reasonably mimics human fixation (AUROC of 0.78 for anticipatory salient patch prediction) and in Fig. 3 (left), the RNN-based selection (blue line) noticeably outperforms random selection (orange line) and slightly under-performs human attention (gray line).

We perform a similar experiment with pedestrian tracking (MOT17 dataset [25]). We combine the RNN-based anticipatory patch selection model with a pre-trained transformer-based tracking model [29]. Foreground object segmentation masks are used as ground truth for training the RNN. The MOTA and MOTP metrics [25] are used for evaluation. We see improvement in MOTA (higher better) and MOTP (lower better) (Fig. 3 (right)) when using learned selection over random selection.

### 3.2    Validation of Architectural Design Choices for Detection

We want to understand how intelligent patch selection and model compression affect object detection. We vary input type (RGB or Bayer), the percent of pixels sensed, the compression of the features, and the type and level of noise. Experiments are conducted on a subset of the MS-COCO validation dataset [23] using mean average precision (mAP, higher is better) as the evaluation metric. Because MS-COCO consists of static images, we use ground truth segmentation maps as a proxy for the patch selection. Results appear in Table 1 (left). We observe: 1) Bayer pixels with reduced features hurts detection accuracy but still achieves high performance ($> 40$ mAP), suggesting a trade-off between efficiency and detection performance; 2) focusing on salient patches aids detection, improving the mAP by 2-4 points; and 3) the system is relatively robust to small amounts of noise, decreasing mAP by only ~3 points.

We hypothesize the loss of performance from model compression is due to: 1) information loss because Bayer pixels discard one-third of the data; 2) error due to domain shift from distilling the RGB model to work with Bayer pixels, and 3) a mismatch between the architectural bias of the feature extractor and structure of Bayer pixels.

**Table 1.** Understanding system performance on detection (left) and MOT (right) as input type (RGB or Bayer), % of pixels sensed, compression of features, and type/level of noise are varied.

| Input | Percent Sensed | Noise Type | Feature Size per Patch | mAP |
|---|---|---|---|---|
| RGB | 100% | None | 768 | 49.0 |
| Bayer | 100% | None | 768 | 43.8 |
| RGB | 100% | None | 240 | 41.4 |
| RGB | 30% | None | 240 | 51.6 |
| Bayer | 30% | None | 240 | 47.6 |
| Bayer | 30% | Gaussian Std Dev = 10 | 240 | 44.8 |
| Bayer | 30% | Poison $\lambda = 1$ | 240 | 45.0 |

| Setting | MOTA | MOTP | Saliency AUROC | Detection Precision | Detection Recall |
|---|---|---|---|---|---|
| Baseline (Full RGB Image) | 0.387 | 0.224 | N/A | 0.917 | 0.464 |
| RGB (30% Sensed, 768-Dim. Features, No Noise) | 0.375 | 0.199 | 0.947 | 0.829 | 0.531 |
| Bayer (30% Sensed, 240-Dim. Features, No Noise) | 0.253 | 0.201 | 0.930 | 0.900 | 0.370 |
| Bayer (30% Sensed, 240-Dim. Features, Gaussian, Std Dev = 10) | 0.211 | 0.206 | 0.928 | 0.862 | 0.350 |
| Bayer (30% Sensed, 240-Dim. Features, Poisson $\lambda = 1$) | 0.234 | 0.204 | 0.930 | 0.881 | 0.362 |

### 3.3    Validation of Architectural Design Choices for Multi-Object Tracking

Extending the previous experiment, we explore how the full system (including the learned patch selection) performs on MOT as input type, percent of pixels sensed, compression of the features, and type and level of noise are varied. Experiments are conducted on a subset of the BDD100K MOTS dataset [38]. This challenging dataset for MOT is captured from the perspective of moving vehicles in crowded environments. We process 32 high-resolution (1280x720) videos at 30FPS. Evaluation metrics include MOTA (higher better), MOTP (lower better), AUROC for anticipatory patch-level saliency prediction (1.0 is perfect), and detection precision and recall. From Table 1 (right), we see little performance loss from sensing partial images vs full images in most metrics. The anticipatory patch selection always achieves over 0.92 AUROC, suggesting the anticipatory sensing mechanism is not the cause of low MOT performance. We expect MOT performance could improve substantially with state-of-the-art detection and tracking models. For reasons discussed in the previous section, using Bayer pixels and shrinking the in-pixel feature size lowers the tracking performance (by ~12 points in terms of MOTA), suggesting there is a trade-off between faster processing and higher performance. I.e., the partially-sensed RGB model achieves performance on par with the full model with 3.3X bandwidth reduction, and if further bandwidth reduction (10X) is needed, then the Bayer model can be used with some loss in performance.

### 3.4   Analysis of Anticipatory Sensing Mechanism for Multi-Object Tracking

We perform ablations to understand each component of the anticipatory patch selection mechanism (Table 2), using RGB images without model compression or added noise. Random patch selection appears to be a strong baseline (0.338 MOTA), suggesting that our model is relatively robust to noisy selection of patches. Using any of the three score functions (RNN-based saliency prediction, object detection uncertainty, and change in tracking uncertainty) improves MOTA by 4-5 points and saliency AUROC by up to 40-45 points, suggesting that each score function is useful on its own. The RNN component results in the highest saliency AUROC (0.948) of any individual component. Unfortunately, we find ensembling the score functions with equal weights does not noticeably improve metrics. To improve the anticipatory sensing mechanism, we may want to dynamically learn to weigh the score components on a per-frame basis.

**Table 2.** Understanding the effects of each component of the anticipatory patch selection mechanism (30% of patches sensed on RGB images without model compression or added noise)

| Score Function | MOTA | MOTP | Saliency AUROC | Detection Precision | Detection Recall |
|---|---|---|---|---|---|
| Random | 0.338 | 0.205 | 0.500 | 0.765 | 0.559 |
| RNN | 0.373 | 0.119 | 0.948 | 0.827 | 0.532 |
| Detection Confidence | 0.375 | 0.119 | 0.903 | 0.830 | 0.531 |
| Tracking Uncertainty | 0.373 | 0.199 | 0.900 | 0.826 | 0.533 |
| All | 0.375 | 0.199 | 0.947 | 0.829 | 0.531 |

### 3.5   Understanding Errors in Multi-Object Tracking

We look across the errors the system makes (Table 3). We measure: how frequently an object track ID switches, the number of frames (on average) until each object is detected for the first time, what percent of objects are detected at least once, overall instance-level recall of object detection, the number of frames (on average) until the model attends to the patches containing each object, and what percent of objects are sensed at least once based on the anticipatory attention mechanism. We analyze results by size of the object (small: $< 32 \times 32$ pixels, medium: $32 \times 32$ to $96 \times 96$ pixels, large $> 96 \times 96$ pixels) and whether the object is occluded or truncated. The system performs well on medium and large objects but struggles on small objects. The system is not overly sensitive to occluded or truncated objects. The system does well anticipating object location (attention is typically placed on the object within 1-2 frames). Objects often appear from the horizon or the periphery, so it takes several frames (mean of 4-5 with high std dev for RGB setting) before most objects are large enough to be initially detected. Once an object is detected, association is decent (1-2 track ID switches per object on average for RGB). Compression (Bayer + 240-dim features) leads to longer time-to-detect, more ID switches, and lower detection recall, which explains the loss in tracking performance.

### 3.6   Energy and Latency Benefits

Our system (anticipatory sensing + model compression) uses 103 GFLOPs/frame, 19.3M model parameters, 113 W of board power (NVIDIA A6000), and 61 ms/frame processing time. An equivalent system without anticipatory sensing and compression uses 1022 GFLOPs/frame, 118M model parameters, 173 W of board power (A6000), and 180 ms/frame processing time. We compute the Energy Delay Product (EDP), which summarizes energy consumption and execution time as a single value. Looking at the ideal EDP (assumes no loss of power to data movement) and measured EDP (uses measured board power), we find energy-latency savings of 15-30X (Table 4).

**Table 3.** Examining mistakes made by the system

RGB - 30% Sensed - 768-Dim Features - No Noise

| Condition | Mean # ID Switches | Time-to-Detect | Percent Detected | Recall Detected | Time-to-Anticipate | Percent Anticipated |
|---|---|---|---|---|---|---|
| All | $1.6 \pm 2.2$ | $4.3 \pm 8.1$ | 0.75 | 0.54 | $0.11 \pm 0.58$ | 0.99 |
| Small Objects | $1.4 \pm 2.0$ | $5.4 \pm 9.5$ | 0.49 | 0.29 | $0.13 \pm 0.66$ | 0.98 |
| Medium Objects | $1.7 \pm 2.4$ | $4.1 \pm 8.0$ | 0.83 | 0.60 | $0.09 \pm 0.49$ | 1.00 |
| Large Objects | $1.5 \pm 2.0$ | $3.9 \pm 6.7$ | 0.97 | 0.86 | $0.15 \pm 0.66$ | 1.00 |
| Occluded | $1.7 \pm 2.3$ | $4.7 \pm 8.2$ | 0.71 | 0.44 | $0.12 \pm 0.60$ | 0.99 |
| Truncated | $0.6 \pm 1.2$ | $1.7 \pm 5.0$ | 0.72 | 0.74 | $0.11 \pm 0.63$ | 0.98 |

Bayer - 30% Sensed - 240-Dim Features - No Noise

| Condition | Mean # ID Switches | Time-to-Detect | Percent Detected | Recall Detected | Time-to-Anticipate | Percent Anticipated |
|---|---|---|---|---|---|---|
| All | $3.2 \pm 4.7$ | $7.0 \pm 10.9$ | 0.57 | 0.38 | $0.23 \pm 1.05$ | 0.99 |
| Small Objects | $2.5 \pm 4.0$ | $8.8 \pm 12.7$ | 0.27 | 0.14 | $0.23 \pm 1.16$ | 0.98 |
| Medium Objects | $3.0 \pm 4.1$ | $6.9 \pm 10.6$ | 0.63 | 0.43 | $0.21 \pm 1.01$ | 0.99 |
| Large Objects | $4.0 \pm 5.9$ | $6.3 \pm 10.4$ | 0.91 | 0.67 | $0.28 \pm 0.96$ | 1.00 |
| Occluded | $3.1 \pm 4.1$ | $7.7 \pm 11.3$ | 0.52 | 0.28 | $0.23 \pm 1.06$ | 0.99 |
| Truncated | $1.5 \pm 3.4$ | $3.2 \pm 9.8$ | 0.49 | 0.49 | $0.17 \pm 0.69$ | 0.98 |

**Table 4.** Comparing EDP of proposed vs baseline system in ideal and measured scenarios

| **Ideal Scenario:**<br>Assuming equal GFLOPS/W:<br>Power Ratio (PR) =<br>(1022 GFLOPs/frame) / (103 GFLOPs/frame) ≈ 9.9X<br>Ideal EDP =<br>PR * ((180 msec/frame) / (61 msec/frame)) ≈ 29.3X | **Measured Scenario:**<br>Assuming 30 W board idle power:<br>Power = Board Power * Framerate<br>Baseline Power (BP) =<br>(173.2 W - 30W) * (0.180 s/frame) ≈ 25.8Ws/frame<br>Proposed system power (PP) =<br>(113.4W - 30W) * (0.061 s/frame) ≈ 5.1Ws/frame<br>Power Ratio (PR) = BP / PP ≈ 5.1X<br>Measured EDP =<br>PR * ((180 msec/frame) / (61 msec/frame)) ≈ 15.3X |
|---|---|

## 4   Conclusions

We presented a system that integrates 1) an architecture for dynamic data-driven sensing at the pixel-level with 2) intelligent processing for computer vision in a feedback control loop. This study highlighted some challenges of data-driven pixel-level control: how to leverage DDDAS for anticipatory attention, how intelligent pixel-level control leads to less complex ML models, and how analog design choices affect predictive performance. Our solution, based on in-pixel processing and anticipatory sensing, led to a 10X reduction in bandwidth and a ~15-30X improvement in energy-latency savings compared to traditional hardware. We demonstrated the effectiveness of ML-based anticipatory patch selection, highlighted accuracy vs efficiency trade-offs, and suggested avenues for improvement (e.g., better handling of small objects and Bayer pixels). Intelligent, dynamic anticipatory sensing combined with novel energy-efficient hardware is useful beyond electro-optic scene understanding. The framework is compatible with other hardware (e.g., pixel-level exposure control), modalities (e.g., RF sensing), domains (e.g., situational awareness via satellite-based sensing), and applications where data reduction is important. The framework for partial adaptive sensing with flexible control can also be extended to manage and reduce data transmission between nodes in a sensor network with limited communication.

**Acknowledgments** This research was, in part, funded by the Defense Advanced Research Projects Agency (DARPA) under Agreement No. HR00112190119. The views, opinions, and/or findings expressed are those of the author(s) and should not be interpreted as representing the official views or policies of the Department of Defense or the U.S. Government.